\pdfoutput=1

\documentclass[11pt]{article}

\usepackage{acl2023}

\usepackage{times}
\usepackage{latexsym}

\usepackage[T1]{fontenc}

\usepackage[utf8]{inputenc}

\usepackage{microtype}

\usepackage{inconsolata}

\usepackage{CJKutf8}
\usepackage{booktabs}
\usepackage{tabularx}
\usepackage{bbding}
\usepackage{graphicx}
\usepackage{bbding}

\title{A Corpus for Named Entity Recognition in Chinese Novels with Multi-genres}

    \author{Hanjie Zhao, Jinge Xie, Yuchen Yan, Yuxiang Jia\thanks{\enspace Corresponding author}, Yawen Ye, Hongying Zan 
\\
       School of Computer and Artificial Intelligence, Zhengzhou University, Zhengzhou, China  \\     
       \texttt{hjzhao\_zzu@163.com,}
        \texttt{\{ieyxjia,iehyzan\}@zzu.edu.cn}
}

\begin{document}
	\begin{CJK*}{UTF8}{gbsn}

\maketitle

\begin{abstract}
Entities like person, location, organization are important for literary text analysis. The lack of annotated data hinders the progress of named entity recognition (NER) in literary domain. To promote the research of literary NER, we build the largest multi-genre literary NER corpus containing 263,135 entities in 105,851 sentences from 260 online Chinese novels spanning 13 different genres. Based on the corpus, we investigate characteristics of entities from different genres. We propose several baseline NER models and conduct cross-genre and cross-domain experiments. Experimental results show that genre difference significantly impact NER performance though not as much as domain difference like literary domain and news domain. Compared with NER in news domain, literary NER still needs much improvement and the Out-of-Vocabulary (OOV) problem is more challenging due to the high variety of entities in literary works. Our data and models are open-sourced at \url{https://github.com/hjzhao73/MultiGenre-ChineseNovel}.
			
\end{abstract}

\section{Introduction}

Named Entity Recognition (NER)\citep{li2020survey} is a crucial task in natural language processing with various applications including information retrieval, text summarization, question answering, machine translation, and knowledge graph. Its objective is to identify specific entities such as person, location, and organization from text. Although great progress has been made in news domain and some vertical domains, NER research in literary domain has been limited due to the lack of annotated data ~\citep{jockers2013macroanalysis}.

To promote the research of literary NER, we build the first NER corpus of online Chinese novels with multi-genres, which contains 260 novels from 13 genres, totaling 105,851 sentences, 5,379,749 Chinese characters, 263,135 entities and 24,458 unique entities of three types person, location and organization. Based on the corpus, we analyze characteristics of entities from different genres. For literary NER, we compare different baseline models and conduct cross-genre and cross-domain experiments. We find that genre difference significantly impact NER performance though not as much as domain difference like literary domain and news domain.

The main contributions of this paper are as follows:
\begin{enumerate}
\item[$\bullet$] We build the first large-scale corpus of online Chinese novels with multi-genres for literary NER and we will release it to the public later.
\item[$\bullet$] We analyze characteristics of entities from different genres and carry out cross-genre and cross-domain experiments for literary NER.
\end{enumerate}	

		\section{Related Work}
		Currently, there is relatively little research on NER in the literary field due to the diverse types of entities and significant differences in naming styles and background knowledge\citep{labatut2019extraction}. Establishing a general NER model for the literary field is challenging, and the lack of large-scale NER datasets limits the development of NER research in this domain\citep{augenstein2017generalisation}.
			
Several previous studies have proposed different approaches and built different corpora for named entity recognition in literary works.
  \citet{vala-etal-2015-mr} introduce a graph-based pipeline model specifically for character recognition.
  \citet{brooke-etal-2016-bootstrapped} propose the LitNER model, which utilizes the bootstrap method.
   \citet{bamman-etal-2019-annotated} build LitBank corpus, annotating named entities in 100 English novels and trained an NER model tailored to the literary field.  \citet{dekker2019evaluating} conduct an evaluation of natural language processing tools to extract characters and build social networks from novels. For Chinese literary NER, \citet{xu2017discourse} construct a dataset for NER and relationship extraction from essays.
  In addition, we
  \cite{jia-etal-2021-rong} create a named entity dataset using Jin Yong's novels and develop a named entity recognition model that incorporates document-level information. The overview statistics of above datasets are shown in Table \ref{tab:dataset of literary}.
  
  However, for Chinese novels, the existing NER corpus is limited in scale and genre. To build a larger-scale multi-genre NER corpus is necessary to enhance further research of literacy NER in Chinese novels.

		\begin{table*}[htbp]
			\centering
			\caption{Literary NER datasets.}
			\label{tab:dataset of literary}
			\begin{tabular}{lllll}
				\toprule
				Dataset & Language  & Tags & Release-Year & Size \\
				\midrule

				LitBank~\cite{bamman-etal-2019-annotated} & English  & 6 & 2019 & 200,000 words\\
				
				Chinese-Literature-NER~\cite{xu2017discourse} & Chinese & 7 & 2017 & 28,897 sentences \\
				
				JinYong~\cite{jia-etal-2021-rong} & Chinese  & 4 & 2021 & 21,927 sentences\\
				\bottomrule
			\end{tabular}
		\end{table*}

		\section{Corpus Construction}
From Qidian Chinese website\footnote{https://www.qidian.com/all/}, we collect novels of 13 different genres, including Xianxia(仙侠), Sport(体育), Military(军事), History(历史), Fantasy(奇幻), Suspense(悬疑), Wuxia(武侠), Game(游戏), Xuanhuan(玄幻), Reality(现实), Sci-Fi(科幻), Urban(都市), and Light Novel(轻小说). For each genre, we crawl the top 20 works from the genre's collection list (as of 2021) and annotate the first 10 chapters of each selected work. All annotated chapters are publicly accessible.

\subsection{Entity Annotation Guidelines}

Considering the characteristics of online Chinese novels, we focus on three entity types,
person (PER), 
location (LOC), and organization (ORG). We follow the entity annotation guidelines of	ACE~\cite{linguistic2005ace}. In addition, (1) We omit single-character entities due to their high ambiguity. (2) We do not annotate nested entities and only annotate the longest one. (3) An entity is composed of head nouns without quantifiers, pronouns and adjective modifiers, etc. (4) An entity must refer to a specific entity in the novel.
 
		\subsubsection{Person}
		Person entities in texts can be represented by various features. For instance, real names of characters such as ``高远''(Gao Yuan) can serve as entities. Additionally, a character's occupation, like ``医生''(doctor), or family relationship, such as ``父亲''(father), can also be indication of entity. Furthermore, a general term like ``小男孩''(little boy) can be used to represent a person entity. Relationships between characters can be denoted by a set of characters, like ``父子''(father and son). Nicknames, such as ``菜鸟''(novice), can also indicate person entities. In the case of deceased individuals or human remains, they could be recorded as person entities, like ``丧尸''(zombie). Even nouns referring to animals or non-human entities, such as ``兽人''(beastman) or ``冰蚕''(ice silkworm), can be used to describe person entities in some genres.
		\subsubsection{Location}
		Location entities typically refer to entities that denote a specific location, such as countries (e.g. 西域,Western Regions) that do not necessarily have a political status, cities (e.g. 羊城,Sheep City), and natural features such as mountains and rivers (e.g. 泰山,Mount Tai).  In Chinese novels locations mostly refer to where the story takes place (e.g. 餐馆,restaurants,训练场,training grounds,小镇, small town).
		\subsubsection{Organization}
		The named entities in the corpus include a range of organizations, such as government agencies (e.g. 组织部,organizational departments), political parties (e.g. 共产党,Communist Party), corporations, universities, high schools, and religious organizations (e.g. 光明圣教,Bright Holy Church). Notably, a substantial portion of the organizational entities in the Chinese novel corpus are fictional, created based on the authors' imagination and settings (e.g. 皇家魔法学院,the Royal School of Magic).

		\subsection{Inter-annotator Agreement}

		To ensure consistent and high-quality annotation, we adopt a multi-round iterative approach. Two annotators simultaneously annotate each novel, cross-check and review each other's work, guaranteeing reliable results. The annotation process consisted of two stages: experimental and formal annotation. In the experimental stage, we use the LTP~\cite{che2020n} named entity recognition tool  to pre-annotate the novels' text, gain familiarity with the corpus and improve the annotation guidelines. In the formal annotation stage, one annotator initially annotates the text, which is then verified by a second annotator to resolve any inconsistencies. The final results are confirmed by the first annotator. This process involves seven annotators and is completed in 70 days.
		
		We assess annotation consistency using the F1 score as the evaluation metric. Results show a micro-averaged F1 score (MicroF1) of 92.15\% and a Micro-averaged F1 score (MacroF1) of 88.62\%, indicating high reliability of the dataset~\cite{artstein-poesio-2008-survey}. The consistency varies across entity types, with person entities demonstrating higher consistency compared to organization and location entities. The complex structures of organization and location entities pose challenges in identifying their boundaries. Detailed values are given in Table~\ref{tab:consistence}.

		\begin{table}[htbp]
			\centering
			\caption{Inter-annotator agreement.}
			\begin{tabular}{ll}
				\toprule
                Entity & F1-score(\%)\\
				\midrule
                PER&93.65\\
                LOC& 90.66\\
                ORG&81.56 \\
                MicroF1&92.15\\
                MacroF1&88.62\\
				\bottomrule
			\end{tabular}%
			\label{tab:consistence}%
		\end{table}
		\subsection{Corpus Analysis}
        Table~\ref{tab:datasets} presents the statistical information of the dataset, which consists of 260 novels covering 13 genres. The dataset includes a total of 105,851 sentences and 263,135 named entities.
  	\begin{table}[htbp]
			\centering
			\caption{Statistics of corpus.}
			\begin{tabular}{llll}
				\toprule
				Entity & Count & Distinct &  Avg.Length \\
				\midrule
				PER & 197,597 & 17,013 & 3.64 \\
				LOC & 45,094 & 4,641  & 3.60 \\
				ORG& 20,444 & 2,804  & 4.87 \\
				Total & 263,135 & 24,458 &3.73 \\
				\bottomrule
			\end{tabular}
			\label{tab:datasets}
		\end{table}

        In general, the dominant type of named entities in novels is person, highlighting the focus on protagonists. Locations constitute the second largest category, serving as the backdrop for storylines and descriptive environments. On the other hand, named entities pertaining to organizations are relatively rare. Additionally, person and location entities tend to be shorter, with an average of 3.64 and 3.60 Chinese characters, respectively, while organization entities tend to be longer, averaging 4.87 Chinese characters. The specific statistics are shown in the Table ~\ref{tab:genre_statistics}, where the largest proportion and average length of each entity type are in bold.		
		
		\begin{table*}[htbp]
			\centering
			
			\caption{Statistics of entities in Chinese novels of various genres. DC represents distinct count, while DR refers to distinct ratio.}
			\label{tab:genre_statistics}
			\begin{tabular}{ccccccc}
				\toprule
				 Genre & Entity &  Count &  Ratio(\%) &  DC &   DR(\%) &  Avg.Length \\ 
				
				\midrule

                            & PER & 18329 & 74.39 & 1817 & 62.89 & 3.18 \\
                    Xianxia & LOC &  4471 & 18.15 & 818 & 28.31 & 2.95 \\
                            & ORG & 1839 & 7.46 & 254 & 8.79 & 3.19 \\ \hline
    
                     & PER & 16641 & 70.12 & 1964 & 64.04 & \bf 4.32 \\
                    Sport & LOC & 3433 & 14.47 & 564 & 18.39 & \bf3.46 \\
                     & ORG &  3658 & \bf 15.41 &  539 & 17.57 & \bf 4.82 \\ \hline

                     & PER & 16365 & 74.03 & 1540 & 56.08 & 3.10 \\
                    Military & LOC & 3589 & 16.23 & 689 & 25.09 & 2.84 \\
                     & ORG & 2153 & 9.74 & 517 & \bf 18.83 & 3.99 \\ \hline

                     & PER &  19925 & \bf 80.61 &  2447 & \bf 70.05 & 3.07 \\
                    History & LOC & 3822 & 15.46 &  820 & 23.48 & 2.54 \\ 
                     & ORG & 970 & 3.92 & 226 & 6.47 & 2.94 \\ \hline
                    
                     & PER & 13617 & 73.36 & 1717 & 63.95 & 3.70 \\
                    Fantasy & LOC & 3327 & 17.92 & 649 & 24.17 & 3.06 \\
                     & ORG & 1618 & 8.72 & 319 & 11.88 & 4.01 \\ \hline
                     
                     & PER & 12897 & 77.00 & 1479 & 61.65 & 3.18 \\
                    Suspense & LOC & 3127 & 18.67 & 708 & \bf 29.51 & 2.94 \\
                     & ORG & 725 & 4.33 & 212 & 8.84 & 3.37 \\ \hline
                    
                     & PER & 17482 & 75.64 & 1976 & 66.60 & 3.18 \\
                    Wuxia & LOC & 4046 & 17.51 & 742 & 25.01 & 2.70 \\
                     & ORG & 1585 & 6.86 & 249 & 8.39 & 3.09 \\ \hline
                    
                     & PER & 14758 & 71.97 & 1805 & 64.44 & 3.32 \\
                    Game & LOC & 4069 & \bf 19.84 & 663 & 23.67 & 2.87 \\
                     & ORG & 1679 & 8.19 & 333 & 11.89 & 3.70 \\ \hline
                    
                     & PER & 17189 & 76.51 & 1547 & 62.94 & 3.29 \\
                    Xuanhuan & LOC & 3846 & 17.12 & 673 & 27.38 & 2.91 \\
                     & ORG & 1432 & 6.37 & 238 & 9.68 & 3.46 \\ \hline
                     
                     & PER & 15280 & 76.75 & 1570 & 62.50 & 3.22 \\
                    Reality & LOC & 3163 & 15.89 & 647 & 25.76 & 2.99 \\
                     & ORG & 1467 & 7.37 & 295 & 11.74 & 3.69 \\ \hline
                     
                     & PER & 13993 & 73.55 & 1555 & 57.04 & 3.53 \\
                    Sci-Fi & LOC & 3702 & 19.46 & 770 & 28.25 &  3.35 \\
                     & ORG & 1329 & 6.99 & 401 & 14.71 & 4.22 \\ \hline
                     
                     & PER & 6039 & 71.67 & 835 & 57.63 & 3.14 \\
                    Urban & LOC & 1537 & 18.24 & 386 & 26.64 & 2.83 \\
                     & ORG & 850 & 10.09 & 228 & 15.73 & 3.90 \\ \hline
                     
                     & PER & 15082 & 78.62 & 1480 & 64.94 & 3.36 \\
                    Light Novel & LOC & 2962 & 15.44 & 578 & 25.36 & 2.98 \\
                     & ORG & 1139 & 5.94 & 221 & 9.70 & 3.64 \\
				 
				\bottomrule
				
			\end{tabular}
		\end{table*}

        Furthermore, we perform genre-specific statistics and identify distinct characteristics in high-frequency person, location, and organization entities among different literary genres. In Table~\ref{tab:genre}, we highlight several genres that exemplify these distinctive characteristics.
	\begin{table*}[htbp]
		\centering
		\caption{Common entities in Chinese novels of different genres.}
		\label{tab:genre}
		\begin{tabularx}{\textwidth}{llX}
			\toprule
			 Genre & Type &  High-frequency entities \\ 
			\midrule
			& PER   & 球迷(Fan),教练 (Coach),于指导(Guidance Yu ),职业球员(Professional player) \\ 
			Sport & LOC   & 中国(China),美国(USA),英格兰(England),西雅图(Seattle)\\ 
			& ORG   &NBA,青年队(youth teams),森林队(forest teams)	\\ 
			\hline		
			
			& PER   & 太宰(Taizai),崇祯(Chongzhen),大魏天子(Emperor of Wei),刘总管(General Manager Liu)  \\ 
			History & LOC   & 秦国(Qin State),京城(Capital),汴梁(Bianliang),宜城(Yicheng) \\ 
			& ORG   &锦衣卫(Jinyiwei),中书省(Zhongshu Province),东宫卫队(the Eastern Palace Guard),豫山书院(Yushan Academy)\\ 
			\hline
			
			& PER   &老法师(Old Mage),女巫(Witch),黑衣武士(Black-clad Warrior),魔法师(Magician)\\ 
			Fantasy & LOC   &  城堡(Castle),鲜花镇(Flower Town),乌山镇(Wushan Town),荆棘岭(Thornridge),圣域(Sanctum)\\ 
			& ORG   & 神盾局(S.H.I.E.L.D.),魔法学院(Academy of Magic),死局帮(Deadlock Gang)\\ 
			\hline

			& PER   & 玄幽道人(Xuanyou Taoist),独孤败天(Dugu Baitian),刘三刀(Liu Sandao),司徒傲月(Situ Aoyue) \\ 
			Wuxia   & LOC   & 蜀山(Shu Mountain),华山(Hua Mountain),通州(Tongzhou),中原(Central Plains) \\ 
			& ORG   & 飞鹰帮(Feiying Gang),画剑派(Huajian Sect),李家(Li Family)\\ 
			\hline
			
			& PER   &青年导游(Youth Tour Guide),赢胖子(Fatty Ying),王秘书(Secretary Wang),副经理(Deputy Manager)\\ 
			Urban   & LOC   & 商场(Shopping malls),网吧(internet cafes),中云市(Zhongyun City),办公室(offices),江南(Jiangnan)\\ 
			& ORG   &医院(hospitals),学府(academic institutions),大学(universities),战争学府(war academies) \\ 
			\bottomrule

		\end{tabularx}
	\end{table*}
 
		For sport genre, high-frequency location entities are typically real-world places such as continents, countries, and cities, while high-frequency organization entities include universities, teams, and leagues. For history genre, high-frequency location entities refer to ancient countries or cities, and high-frequency organization entities are ancient government institutions such as the ``锦衣卫''(Jinyiwei) and ``中书省''(Zhongshu Province). For fantasy and science fiction genres, high-frequency location entities are fictional places like castles, towns, and laboratories, while high-frequency organization entities include fictional organizations like ``神盾局''(S.H.I.E.L.D.),``学院''(academies), and ``联邦''(federations). For urban genre, high-frequency location entities are everyday places, and high-frequency organization entities are companies, hospitals, and universities.

		\section{Literary Named Entity Recognition}
		\subsection{Baseline Models}
		
		The corpus is divided into training, validation, and test sets in an 8:1:1 ratio for this study, which aims to train multiple models for named entity recognition in the literary domain. The F1 scores of various models are compared for the three categories of person, location, and organization, as well as the MicroF1 and MacroF1 scores. 
  
		Table~\ref{tab:different models} demonstrates that BERT-BiLSTM-CRF\citep{devlin-etal-2019-bert,huang2015bidirectional} exhibits the highest values in terms of MicroF1 and MacroF1 metrics, indicating its superior overall performance. The best value on each entity or metric is in bold.
		The recognition performance is best for person, followed by location and organization. 
		This study shows that models using pre-trained model as feature extractor perform the best, while models based only on BiLSTM and CRF\citep{lafferty2001conditional} perform relatively poorly. This highlights the significant enhancement in the overall performance of named entity recognition through the incorporation of pre-trained models.

		\begin{table*}[htbp]
			\centering
			\caption{Comparison of baseline models.}
			\label{tab:different models}
			\begin{tabular}{llllll}
				\toprule
				Model & PER(\%) & LOC(\%) & ORG(\%) & MicroF1(\%) & MacroF1(\%) \\
				\midrule
				BiLSTM-CRF & 78.59 & 64.37 & 52.09 & 74.47 & 65.02 \\
				BERT-CRF &\bf 87.84 &\bf 86.21 & 77.44 & 86.55 & 83.83 \\
				BERT-BiLSTM-CRF & 87.72 & 85.41 & \bf 79.09 &\bf 86.73 &\bf 84.07 \\
				\bottomrule
			\end{tabular}
		\end{table*}
		
        Table~\ref{tab:OOV} provides a summary of the performance of the BERT-BiLSTM-CRF model, with a particular focus on its handling of Out of Vocabulary (OOV) entities. In the test set, the ratio of OOV to in-vocabulary (IV) entities is approximately 1:2, consisting of 1417 OOV entities and 3109 IV entities. The results reveal that the model exhibits declined performance in recognizing OOV entities, particularly struggling in identifying OOV LOC entities, achieving a F1 score of only 31.63\%.
  
        \begin{table*}[htbp]
			\centering
			\caption{OOV vs. IV}
            \label{tab:OOV}
			\begin{tabular}{llllll}
				\toprule
				& PER(\%) & LOC(\%) & ORG(\%) & MicroF1(\%) & MacroF1(\%) \\
				\midrule
				OOV(1417) & 49.70 & 31.63 & 35.27 & 45.07 & 38.67 \\
				IV(3109) & 91.43 & 91.43 & 85.36 & 91.02 & 89.41 \\
				\bottomrule
			\end{tabular}
		\end{table*}

        \subsection{One-model-one-type vs. One-model-all-types}
		To investigate the impact of recognizing multiple entities simultaneously, we train a model with the same parameters as BERT-BiLSTM-CRF while separately training individual entities. As shown in the Table~\ref{tab:individual}, the model that predicts multiple entities simultaneously can allow the model to learn more diverse knowledge, leading to an improvement in the model's recognition performance on single entities. This finding highlights the advantage of incorporating a multi-entity recognition approach, as it enables the model to leverage contextual information and inter-dependencies among entities to enhance its accuracy and effectiveness in named entity recognition tasks.

        \begin{table*}[htbp]
        	\centering
        	\caption{One-model-one-type vs. One-model-all-types.}
        	\label{tab:individual}
        	\begin{tabular}{lccc}
        		\toprule
        		Model&PER(\%)&LOC(\%)&ORG(\%)\\
        		\midrule
        		One-model-one-type& 85.99 & 85.04 & 76.51 \\
        		One-model-all-types & 87.72(+1.73) & 85.41(+0.37) & 79.09(+2.58) \\
        		\bottomrule
        	\end{tabular}
        \end{table*}
		
		\subsection{Cross-genre NER in Novels}
        In this study, we train a BERT-BiLSTM-CRF model for each genre. The corpus is divided into training, validation, and testing sets in an 8:1:1 ratio. We use these models to predict entities in novels from 13 different genres and compare the performance variations across genres using a confusion matrix, where each row represents the predictions made by a specific genre model.
        
\begin{figure}[htbp]
			
			\centering
			\includegraphics[width=\linewidth]{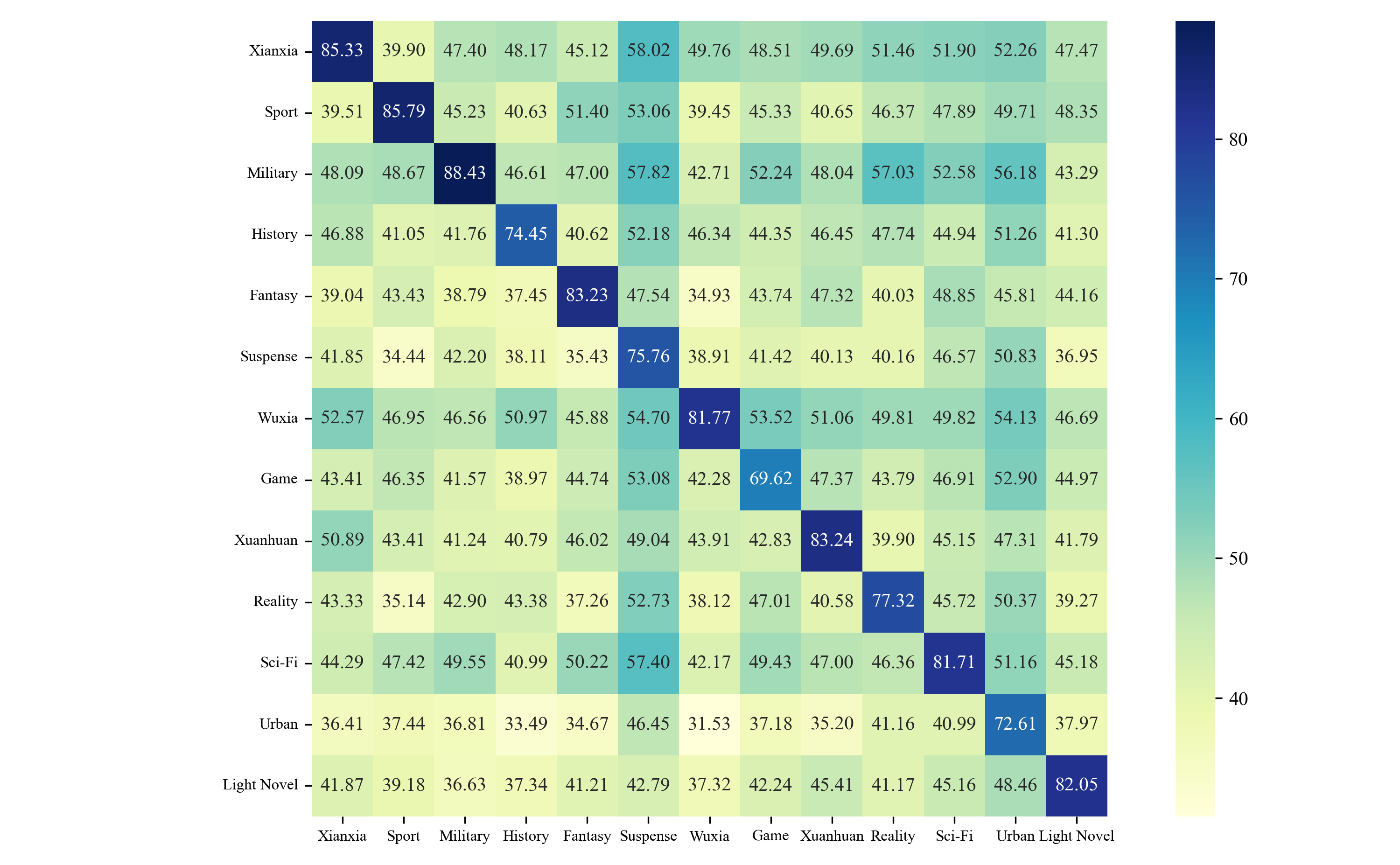}
			\caption{Confusion matrix of MicroF1 for different genres.}
			\label{fig:confusion-micro}
		\end{figure}

        The MicroF1 values of the predictions are shown in Figure ~\ref{fig:confusion-micro}. It is noteworthy that when predicting entities in historical novels, models trained on the Xianxia and Wuxia genres perform well, benefiting from their historical backgrounds. Conversely, the model trained on urban novels set in modern times shows the poorest performance in the historical genre. Through cross-genre experiments, we observe the significant impact of different themes on named entity recognition, even within the same domain of novels.

\begin{figure}[htbp]
	
			\centering
			\includegraphics[width=\linewidth]{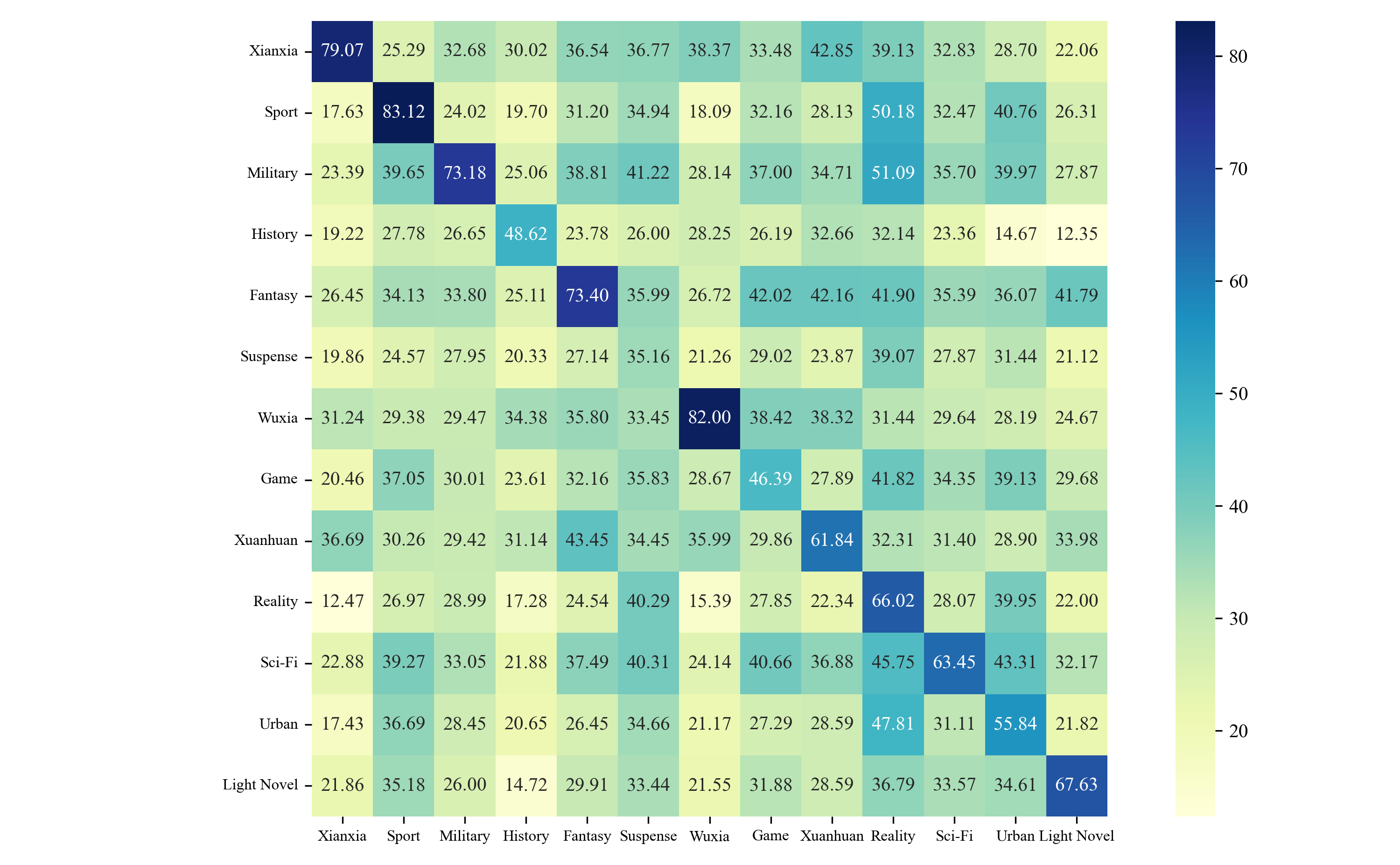}
			\caption{Confusion matrix of ORG-F1 for different genres.}
			\label{fig:confusion-org}
		\end{figure}

        Furthermore, we discover some unexpected results when predicting organization (ORG) entities across genres. As shown in Figure ~\ref{fig:confusion-org}, the model trained on Suspense novels performs even worse when predicting entities in the same genre novels. This could be attributed to the scarcity of organization entities in Suspense novels and the less distinctive thematic features, as evident in Table 4. 
        Additionally, the predictions from the confusion matrix validate certain distributional differences among various genre novels, even for those with similar characteristics. For instance, when predicting organization entities in Reality novels, there is a significant disparity between the Xianxia and Wuxia genres, with a score of 39.13\% for Xianxia and 31.44\% for Wuxia.

        In summary, our study demonstrates the impact of corpus sources on model performance in named entity recognition, showcasing the variations across genres and highlighting the importance of considering genre-specific characteristics in the training and prediction processes.

		\subsection{Cross-domain NER}
		To investigate the degree to which NER depends on domain-specific knowledge, we conduct cross-domain experiments to compare NER performance on different corpora. Specifically, we utilize news articles from People's Daily (Peopledaily) spanning from January to June 1998 as the general domain corpus and compare it with the Chinese novel corpus Qidian. The statistics of the two corpora are shown in Table~\ref{tab:dataset_comparison}. We train NER models on each corpus and compare their performance. As shown in Table~\ref{tab:F1}, NER performance varies significantly across corpora from different domains, indicating its high sensitivity to domain-specific information.

        Furthermore, when we employ the Peopledaily dataset for training our model to predict Chinese novel data, we make an intriguing observation. The F1 score for recognizing ORG entities is remarkably low at 0.47\%, with a recall rate of just 0.0023\%. However, the precision is quite close to that of the other two entity types. We attribute this outcome to the fact that the Peopledaily dataset encompasses numerous political organization entities which are rarely used in online novels.

        \begin{table}[ht]
        \centering
        \caption{Statistics comparison: Qidian vs. Peopledaily}
        \label{tab:dataset_comparison}
        \begin{tabular}{lll}
        \toprule
         & Qidian & Peopledaily \\
        \midrule
        Sentences & 105,851 & 123,882 \\
        Words & 5,379,749 & 11,978,551 \\
        Entities & 263,135 & 323,368 \\
        Unique entities & 24,458 & 43,249 \\
        \bottomrule
        \end{tabular}
        \end{table}

		\begin{table*}[ht]
			\centering
			\caption{Cross-domain NER.}
			\label{tab:F1}
			\begin{tabular}{llllll}
				\toprule
				Domain & PER(\%) & LOC(\%) & ORG(\%) & MicroF1(\%) & MacroF1(\%) \\
				\midrule
				Qidian-Qidian & 87.72 & 85.41 & 79.09 & 86.73 & 84.07 \\
				Qidian-Peopledaily & 44.05 & 46.45 & 13.44 & 38.64 & 34.65 \\
				Peopledaily-Qidian & 49.65 & 23.33 & 0.47  & 42.91 & 24.48 \\
				Peopledaily-Peopledaily & 96.20 & 96.90 & 97.99 & 96.74 & 97.03 \\
				\bottomrule
			\end{tabular}%
			\label{tab:domain_compare}%
		\end{table*}

		\subsection{Case Study}
		Table~\ref{tab:predict} gives two examples for experiments of the section of baseline models. In the first example, even models incorporating pre-trained models wrongly recognize boundaries of person entities, like ``女伯爵''(Countess). Due to the limitations of the training set, the models mislabel the type of entity ``匈牙利''(Hungary).
		In the second example, the person entity ``古河''(Gu He) contains ``河''(river), the frequently occurring suffix of location entity, causing all the models based on pre-trained model to erroneously classify ``古河''(Gu He) as a location entity. These examples fully demonstrate the significant impact of domain and contextual information on named entity recognition.

		\begin{table*}[htbp]
			\centering
			\caption{Case analysis.}
			\label{tab:predict}%
			\begin{tabularx}{\textwidth}{lX}
				\toprule
				 & Predicted Result \\
				\midrule
				Ground Truth(from Game) & 这位\underline{女伯爵}$_{PER}$$\cdots$她都被视为\underline{匈牙利}$_{LOC}$最美的女人之一。 \\ 
				
				BiLSTM-CRF & 这位女\underline{伯爵}$_{PER}$$\cdots$她都被视为\underline{匈牙利}$_{PER}$最美的\underline{女人}$_{PER}$之一。		\\
								
				BERT-CRF &    这位女\underline{伯爵}$_{PER}$$\cdots$她都被视为\underline{匈牙利}$_{PER}$最美的\underline{女人}$_{PER}$之一。		\\		
						
				BERT-BiLSTM-CRF & 这位女\underline{伯爵}$_{PER}$$\cdots$她都被视为\underline{匈牙利}$_{PER}$最美的\underline{女人}$_{PER}$之一。		\\		

				\hline

				Ground Truth(from Xuanhuan) & \underline{古河}$_{PER}$名列\underline{加玛帝国}$_{ORG}$十大强者之一。\\							
				BiLSTM-CRF & \underline{古河}$_{PER}$名列\underline{加玛帝国}$_{ORG}$十大强者之一。\\		
				
				BERT-CRF
				& \underline{古河}$_{LOC}$名列\underline{加玛帝国}$_{ORG}$十大强者之一。\\		
				
				BERT-BiLSTM-CRF 
				& \underline{古河}$_{LOC}$名列\underline{加玛帝国}$_{ORG}$十大强者之一。\\
				
				\bottomrule
				
			\end{tabularx}
		\end{table*}

		\section{Conclusion}
In this paper, we build the largest multi-genre corpus of Chinese novels for literary NER. We describe the annotation guidelines and analyze characteristics and distributions of entities from different genres. We propose several baseline models for literary NER and find that the pre-trained model can significantly improve the performance. Our corpus provides a valuable dataset for cross-genre NER investigation, which shows that genre difference makes obvious decline of performance. 

The cross-domain experiments between literary domain and news domain show that literary NER still needs improvement and domain difference makes much more severe performance drop, reaffirming the necessity of a domain corpus for vertical domain NER. The comparison between one-model-one-type and one-model-all-types NER shows that learning multi-types of entities simultaneously can enhance the entity recognition of each type. 

In the future, we will further study cross-genre and cross-domain problems in literary NER. The OOV problem is more challenging in literary texts, which is another problem we plan to address.

\bibliography{anthology,custom}
\bibliographystyle{acl_natbib}

		\end{CJK*}
\end{document}